\documentclass[10pt,twocolumn,letterpaper]{article}

\usepackage{cvpr}              

\usepackage{algorithm,algpseudocode}
\usepackage{algorithmicx}

\usepackage[dvipsnames]{xcolor}

\definecolor{cvprblue}{rgb}{0.21,0.49,0.74}
\usepackage[pagebackref,breaklinks,colorlinks,citecolor=cvprblue]{hyperref}
\usepackage{bbding}
\title{Portrait Diffusion: Training-free Face Stylization with Chain-of-Painting}

\author{
\vspace{0.2cm}
Jin~Liu$^{1,2}$\quad
Huaibo~Huang$^2$\quad
Chao~Jin$^2$\quad
Ran~He$^{1,2}$\Envelope \\
$^1$School of Information Science and Technology, ShanghaiTech University, China \\ 
$^2$CRIPAC~\&~MAIS, Institute of Automation, Chinese Academy of Sciences, China  \\
\texttt{liujin2@shanghaitech.edu.cn, huaibo.huang@cripac.ia.ac.cn}\\
\texttt{jinchao2024@ia.ac.cn, rhe@nlpr.ia.ac.cn} 
}

\begin{document}

\twocolumn[{%
  \renewcommand\twocolumn[1][]{#1}%
  \maketitle
  \begin{center}
   \centering
   \includegraphics[width=\textwidth]{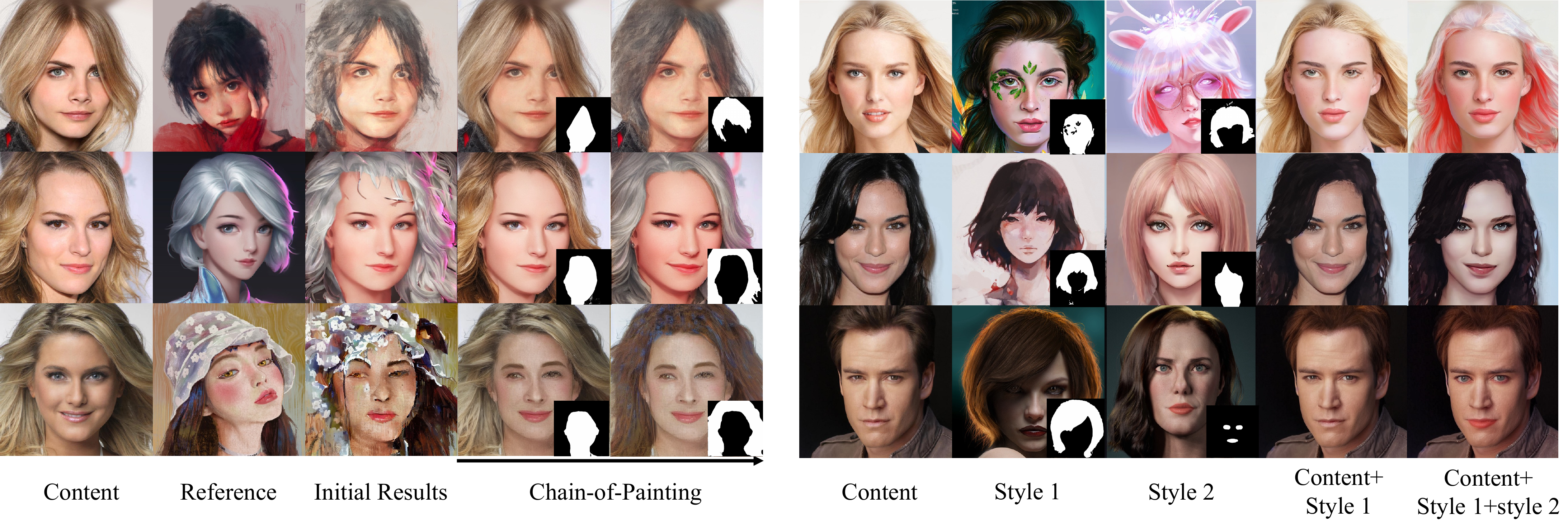}
   \vspace{-15pt}
   \captionof{figure}{Our framework Portrait Diffusion can synthesize images without the need for fine-tuning. Portrait Diffusion can iteratively generate stylized facial images through Chain-of-Painting, (a) using Chain-of-Painting to iteratively redraw poor results, and (b) achieving the transfer of multiple styles by using different style references in Chain-of-Painting.}
   \label{fig:cop}
  \end{center}
 }]

\begin{abstract}

Face stylization refers to the transformation of a face into a specific portrait style. However, current methods require the use of example-based adaptation approaches to fine-tune pre-trained generative models so that they demand lots of time and storage space and fail to achieve detailed style transformation. This paper proposes a training-free face stylization framework, named Portrait Diffusion. This framework leverages off-the-shelf text-to-image diffusion models, eliminating the need for fine-tuning specific examples. Specifically, the content and style images are first inverted into latent codes. Then, during image reconstruction using the corresponding latent code, the content and style features in the attention space are delicately blended through a modified self-attention operation called Style Attention Control. Additionally, a Chain-of-Painting method is proposed for the gradual redrawing of unsatisfactory areas from rough adjustments to fine-tuning. Extensive experiments validate the effectiveness of our Portrait Diffusion method and demonstrate the superiority of Chain-of-Painting in achieving precise face stylization. Code will be released at \url{https://github.com/liujin112/PortraitDiffusion}.

\end{abstract}
    
\section{Introduction}
\label{sec:intro}


The artistic portrait finds wide-ranging applications in our daily lives and creative industries, spanning social media, animation, films, and advertisements. Face stylization seeks to transfer the style in artistic portraits onto a chosen face photo. This process empowers individuals, regardless of their painting skills, to infuse their preferred artistic styles into their photos.


Recent studies \cite{zhu2021mind, zhang2022towards, Yang_2022_Pastiche_Master, chong2022jojogan, zhang2022generalized, zhang2023inversion} have delved into adaptation methods for face stylization. These approaches initially fine-tune a pre-trained generative model, such as StyleGAN \cite{karras2019style} or Stable Diffusion \cite{rombach2022high}, by leveraging a limited set of artistic portrait examples, and subsequently generating a stylized face through the decoding of the source face via the fine-tuned model. Despite their success in transferring diverse portrait styles, maintaining a balance between preserving content and implementing effective style transfer remains a challenge. Additionally, these methods mainly learn a global style representation, thereby encountering challenges in achieving fine-grained control over the style of the generated faces. Furthermore, the computational costs associated with fine-tuning models on each portrait style pose a hurdle to the widespread applicability of these methods across arbitrary portrait styles.




To address these challenges, we introduce \textbf{Portrait Diffusion}, a novel training-free framework for face portrait stylization. Our method employs a pre-trained diffusion model to synthesize a stylized face directly from a given source image and a reference image. Initially, we encode the source and reference images into latent codes using DDIM Inversion \cite{song2020denoising, kim2022diffusionclip}. Subsequently, we produce the stylized image by progressively eliminating the predicted noise from the source latent code. we reconstruct both the source and reference images from their respective latent codes, facilitating the integration of features from both images in each denoising step. To achieve this, we introduce \textbf{Style Attention Control} to replace the self-attention mechanism in the diffusion model, which queries the feature from the reference image instead of itself. Specifically, we employ a dual-branch cross-attention mechanism, with one branch processing the source features and the other the target features. The outputs from these branches are modulated by the fusion ratio through the a Style Guidance scale, allowing for precise control over the stylization intensity. Importantly, Style Attention Control alters only the interaction of features within the self-attention mechanism, permitting the incorporation of style features in a training-free approach.

However, directly querying features in the attention space may lead to the issue of query confusion. This occurs when the features of the reference and source images, despite having different semantic meanings, display similar patterns and colors. To address this, we further introduce a Mask-Prompted Style Attention Control, which leverages two masks, one for the source image and the other for the reference image, to guide Style Attention Control to operate only in areas sharing identical semantic meanings. Additionally, we propose \textbf{Chain-of-Painting}, which augments stylized outcomes through a sequence of facial attribute maps derived from human feedback. Initially, a basic stylization is applied to the source image, followed by iterative refinement of the unsatisfactory areas of the stylized results using mask prompting. In addition to generating higher-quality stylized results, Chain-of-Painting also facilitates the face stylization with multiple style references. To achieve this, we employ varied reference images and masks that denote distinct facial attribute regions at each step of Chain-of-Painting. In this way, we can synthesize a face image that different attributes with different portrait styles.

Our main contributions are summaries as follows:
\begin{itemize}
    \item We propose Portrait Diffusion, a training-free portrait stylization framework, which can flexibly transfer the fine-grained portrait style to a source image.

    \item We propose Style Attention Control to fuse content and style features in the space of self-attention and control the strength of style information via Style Guidance.

    \item  We introduce Chain-of-Painting, a new portrait stylization paradigm which can progressively refine a stylized image or stylize a portrait with multiple style references. 
\end{itemize}




\section{Related works}

\subsection{Few-Shot Face Stylization}

Few-shot face stylization methods leverage models pre-trained on extensive source domain datasets, such as StyleGAN \cite{karras2019style}. Distinct from other image-to-image translation approaches, both supervised \cite{pix2pix2017,wang2018high,zhu2017_toward} and unsupervised approaches \cite{zhu2017unpaired,liu2017unsupervised,kim2019u}, which demand extensive training data. In contrast, few-shot techniques primarily employ GAN Adaptation  \cite{wang2018transferring,noguchi2019image,wang2020minegan,robb2020few,zhao2020leveraging,Yang_2023_One_Shot}. This adaptation method requires only a minimal number of target domain samples (typically not exceeding a few hundred) to fine-tune a pre-trained GAN \cite{karras2019style,karras2020analyzing,karras2020training}. One application of this method is Toonify \cite{pinkney2020resolution}, which fine-tunes StyleGAN using a small collection of cartoon samples. This technique, which involves interpolating between the weights of the fine-tuned and original models, effectively generates convincing cartoon faces. However, the process encounters a significant challenge of overfitting when fine-tuning with limited few-shot samples. To mitigate this issue, research such as \cite{li2020few,ojha2021few} incorporated additional regularization within the latent space. AgileGAN \cite{song2021agilegan} introduced an inversion-consistent transfer learning framework, which markedly diminishes the variance in inversion distribution.

Additionally, \cite{xiao2022appearance} explored an intermediate domain bridging the source and animation domains to minimize the domain gap. DualStyleGAN \cite{Yang_2022_Pastiche_Master} enhanced StyleGAN with extra style paths, facilitating efficient intrinsic and extrinsic style tuning. However, it still requires hundreds of images for fine-tuning, limiting its applicability in data-scarce scenarios. JoJoGAN \cite{chong2022jojogan} proposed an one-shot face stylization approach using GAN style mixing and pixel loss for fine-tuning with a reference image-based paired dataset. Another novel one-shot adaptation method is presented in \cite{zhang2022generalized}, which divides face stylization into style and entity transfer for more natural transformations. Moreover, StyleDomain \cite{alanov2023styledomain} proposed an efficient, lightweight method for domain adaptation by modifying the style vector in StyleSpace. 

\subsection{Diffusion-Based Style Transfer} 

In recent years, diffusion models \cite{sohl2015deep,ho2020denoising,song2020score,song2020denoising,dhariwal2021diffusion} have become preeminent in the generative domain. The advent of a series of text-to-image models \cite{DALLE3,rombach2022high,saharia2022photorealistic} has propelled AI-generated art into widespread popularity. Consequently, a variety of style transfer works based on diffusion models have emerged \cite{kim2022diffusionclip,zhang2023inversion,zhaoegsde2022,preechakulDiffusionAutoencodersMeaningful2022}. Diffusion-based style transfer methods are primarily divided into two categories: energy guidance-based methods and personalization-based methods.

The concept of energy-guided methods \cite{nair2023steered,yu2023freedom,kwon2022diffusion,yang2023zero,Hamazaspyan_2023_CVPR} is derived from classifier guidance \cite{dhariwal2021diffusion}, which employs the gradients of an estimated loss to steer the generation process at each sampling step. Energy-guided methods assess the discrepancy between the output and target at each generative step through a meticulously designed energy function \cite{yu2023freedom,nair2023steered}. Notably, directly utilizing the output of a diffusion model can be time-intensive. Hence, a coarse-grained prediction is often employed as a more efficient alternative. Personalization-based methods necessitate the initial fine-tuning of a pre-trained model via Dreambooth \cite{ruiz2023dreambooth}, Textual Inversion \cite{gal2022image,zhang2023inversion}, or LoRA \cite{hu2021lora}, followed by the decoding of the latent codes of inverted content images using the fine-tuned model. This process bears a striking resemblance to the GAN Adaptation method. However, the inherent stochasticity of diffusion models poses challenges in accurately reconstructing content images. To enhance image reconstruction and content preservation, some studies have integrated Prompt Tuning \cite{mokady2023null,cheng2023general} into a T2I diffusion model. Additionally, several works fine-tuned diffusion models by integrating CLIP \cite{kim2022diffusionclip,Yang_2023_Zero-Shot,wang2023stylediffusion} to learn style references through a text prompt. \cite{cheng2023general} developed a comprehensive framework for diffusion-based image-to-image translation.

Work most akin to Portrait Diffusion is that by Cheng et al. \cite{cheng2023general}. Their Visual Concept Translator (VCT) framework similarly employs attention control during the generation phase. This is achieved by substituting the cross-attention map of the target image with that of the reconstructed source image, thereby enhancing content preservation. In Portrait Diffusion, attention control is primarily utilized for the fusion of style and content features in self-attention mechanism. The VCT framework \cite{cheng2023general} necessitates multi-concept inversion for the style image, rendering it a one-shot method. In contrast, our Portrait Diffusion is a zero-shot method, obviating the need for model fine-tuning or prompt tuning.

\section{Method}

\subsection{Preliminaries}

\noindent
\textbf{Latent Diffusion Models.} 

Latent Diffusion Models (LDMs) \cite{rombach2022high} operate by training a diffusion model within a latent space of a pre-trained autoencoder \cite{esser2021taming}. The encoder $\mathcal{E}$ transforms an image $x$ into a latent code $z_0 = \mathcal{E}(x)$, while the decoder $\mathcal{D}$ is capable of reconstructing the image $x=\mathcal{D}(z_0)$ from $z_0$.

The forward progression of the diffusion model constitutes a Markov chain that incrementally introduces noise into a latent code $z_0$, resulting in a noisy latent code $z_t$ at a given time-step $t$:
\begin{equation}
q\left(z_t \mid z_0\right) := \mathcal{N}\left(z_t ;\sqrt{{\alpha}_t} \boldsymbol{z}_0, \sqrt{1-{\alpha}_t} \boldsymbol{I}\right),
\label{onestepnoised}
\end{equation}
where $\{{\alpha}_t\}^T_{t=0} \in [0,1]$ denotes the predefined diffusion schedule hyperparameters, characterized as a monotonically decreasing sequence that ensures $z_T$ is pure Gaussian noise at the terminal time step $T$.

The reverse process of the diffusion model constitutes an approximated Markov chain. During this process, the noise in $z_T$ is iteratively diminished via a learned Gaussian transition $p_\theta\left(\mathbf{z}_{t-1} \mid \mathbf{z}_t\right)$, culminating in the generation of a noise-free latent code $z_0$. The learned Gaussian transition is formulated as follows:

\begin{equation}
p_\theta\left(\mathbf{z}_{t-1} \mid \mathbf{z}_t\right)=\mathcal{N}\left(\mathbf{z}_{t-1} ; \boldsymbol{\mu}_\theta\left(\mathbf{z}_t, t\right), \sigma_t^2 \boldsymbol{I} \right) ,
\label{eq:ddpm}
\end{equation}
where $\sigma_t$ represents a time-dependent constant. The mean $\boldsymbol{\mu}_\theta$ is parameterized by the following equation: 
\begin{equation}
\boldsymbol{\mu}_\theta\left(\mathbf{x}_t, t\right):=\frac{1}{\sqrt{\alpha_t}}\left(\boldsymbol{x}_t-\frac{1-\alpha_t}{\sqrt{1-\bar{\alpha}_t}} \boldsymbol{\epsilon}_\theta\left(\boldsymbol{x}_t,t\right)\right).
\end{equation}
Here, $\boldsymbol{\epsilon}_\theta\left(\boldsymbol{x}_t, t\right)$ denotes a time-conditioned U-Net \cite{ronneberger2015u}, functioning as a noise estimator designed to predict the noise $\epsilon$. It is trained to minimize the following objectives:
\begin{equation}
L(\theta)=\mathbb{E}_{t,z_0, \epsilon}\left[\left\|\boldsymbol{\epsilon}-\boldsymbol{\epsilon}_\theta\left(\boldsymbol{z}_t,t\right)\right\|^2\right].
\end{equation}

\begin{figure*}[t]
\begin{center}
\includegraphics[width=1\linewidth]{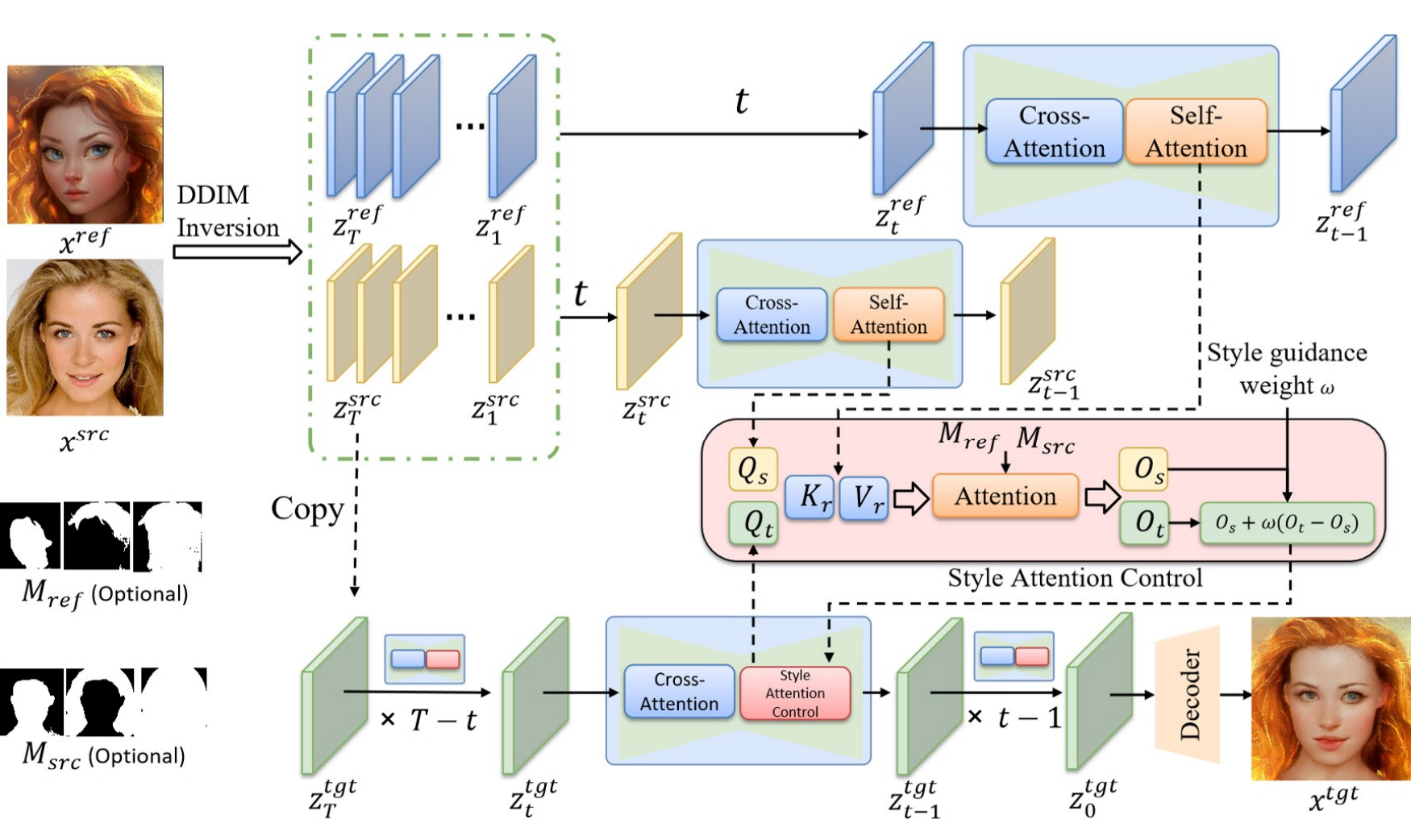}
\end{center}
   \caption{
The overall pipeline of the proposed Portrait Diffusion. The source image and reference image are inverted into latent codes. The generation of the target image initiates from the latent code of the source image. During each denoising step, the original self-attention mechanism is substituted with the newly introduced Style Attention Control. This novel approach effectively integrates the features of the source, reference, and target images in the attention space.}
\label{fig:framework}
\end{figure*}

\noindent
\textbf{DDIM Inversion.}

In this study, we utilize the DDIM inversion method \cite{song2020denoising} to transform real images into latent codes, which are then employed for face stylization. The DDIM inversion is accomplished through the reverse application of the DDIM sampling strategy. Specifically, the DDIM sampling strategy employs a deterministic transition for latent codes $z_t$:
\begin{equation}
    z_{t-1}=\sqrt{\alpha_{t-1}} f_{\theta}\left(z_{t}, t\right)+\sqrt{1-\alpha_{t-1}} \epsilon_{\theta}\left(z_{t}, t\right),
    \label{eq:ddimforward}
\end{equation}
where $f_\theta$ denotes the prediction of $z_0$ from $z_t$ at step $t$. In contrast to the stochastic sampling in DDPM Eq. (\ref{eq:ddpm}), DDIM's approach is deterministic. Utilizing the DDIM process on $z_{t-1}$ to derive $z_{t}$ leads to the inverse DDIM sampling, which is expressed as:
\begin{equation}
    z_{t+1}=\sqrt{{\alpha}_{t+1}} f_{\theta}\left(z_{t}, t, v\right)+\sqrt{1-{\alpha}_{t+1}} \epsilon_{\theta}\left(z_{t}, t\right) .
    \label{eq:ddiminversion}
\end{equation}
Thus, one can convert a clean latent code $z_0$ into a noisy latent code $z_T$ by repetitively applying Eq. (\ref{eq:ddiminversion}) $T$ times, which we describe as:
\begin{equation}
 z_T = \text{DDIMEncode}(z_0).
 \label{eq:ddimencode}
\end{equation}
Given that the inverse DDIM sampling process is deterministic, $z_0$ can be reconstructed by applying Eq. (\ref{eq:ddimforward}) to $z_T$.

\subsection{Portrait Diffusion}
Given a source image $x^{src}$ from the face photo domain and a reference style image $x^{ref}$ from the artistic portrait domain, the objective of Portrait Diffusion is to generate a stylized image $x^{out}$ that retains the structural integrity and semantic layout of $x^{src}$. Owing to the deterministic nature of DDIM inversion, a pre-trained diffusion model is capable of converting any image into latent codes and subsequently reconstructing images from these codes. The central concept of our approach is the fusion of features from the source and reference images during their restoration from the respective latent codes. To facilitate this, our Portrait Diffusion incorporates a cross-attention operation, designated as Style Attention Control, which supersedes the self-attention mechanism within the stable diffusion model. This modification enables the transfer of facial style from the reference to the source image through the querying of analogous semantic features in the reference images.

The overall framework of the proposed Portrait Diffusion is illustrated in Fig.~\ref{fig:framework}. Initially, the source image $x^{src}$ and the reference image $x^{ref}$ are converted to the noised latent space of the diffusion model, $z^{src}_T,z^{ref}_T = \text{DDIMEnc}(\mathcal{E}(x^{src},x^{ref}))$, utilizing the DDIM inversion as Eq. (\ref{eq:ddimencode}). To mitigate the accumulation of errors during reconstruction and ensure more precise reconstruction, we preserve the intermediate latent codes generated throughout the DDIM inversion process. This results in two sequences of latent codes, $\{z^{src}_T\}^T_{t=1}$ and $\{z^{ref}_T\}^T_{t=1}$. During the denoising stage at time step $t$, the features derived from their respective latent codes are fed into the Style Attention Control in the third branch, which commences with the final latent code $z^{src}_T$ of $x^{src}$ for feature fusion. The specifics of this process are expounded in the subsequent section.

\begin{figure*}[!h]
\begin{center}
\includegraphics[width=0.9\linewidth]{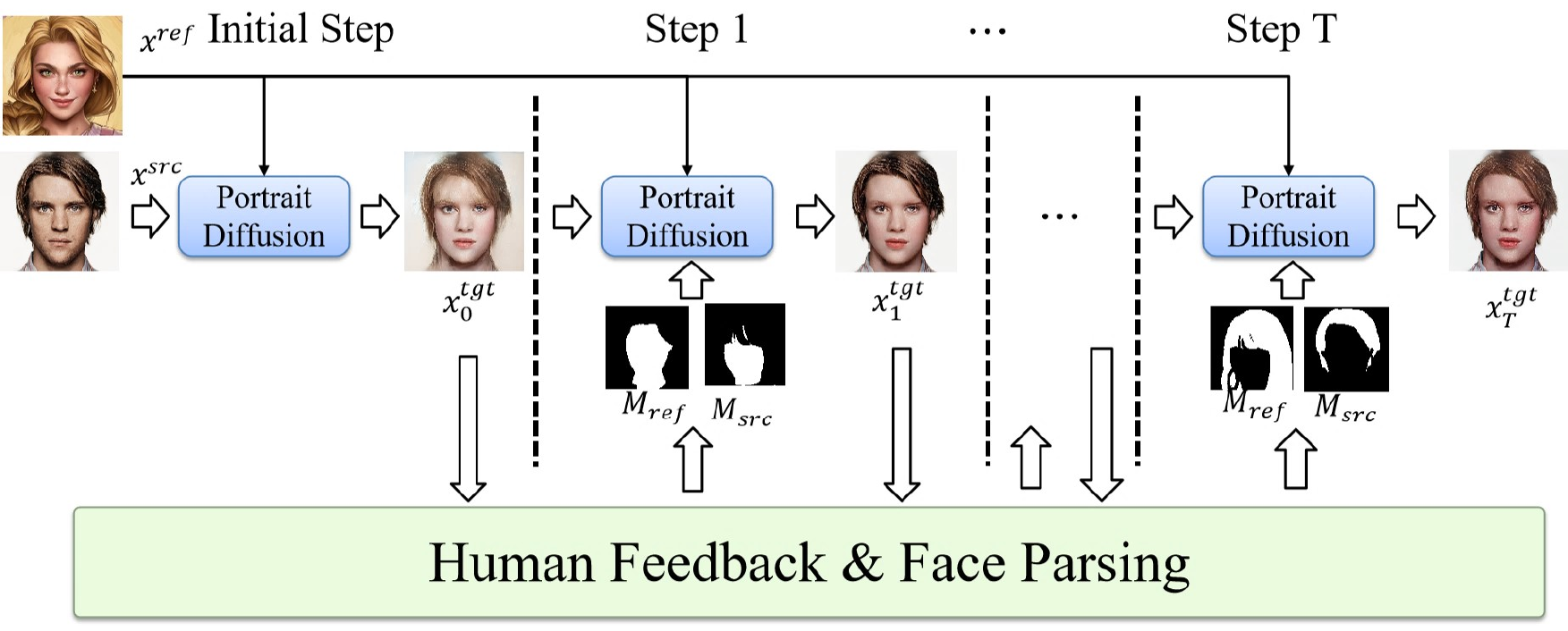}
\end{center}
\vspace{-15pt}
   \caption{ Illustration of the process of Chain-of-Painting. Chain-of-painting involves an iterative process rather than generating a result through a single inference. This method integrates human feedback to progressively synthesize a stylized image in a step-by-step manner.
   }
\label{fig:chain-of-painting}
\end{figure*}

\noindent
\textbf{Style Attention Control.}  
Style Attention Control replaces the original self-attention mechanism in the Stable Diffusion model, implementing a dual-branch cross-attention architecture. This architecture not only utilizes the features derived during the denoising of the stylized image's latent code but also incorporates features from the denoising of the latent codes of source images to query features in the reference image. Specifically, the input for each denoising iteration comprises three latent codes: $(z^{ref}_t, z^{src}_t, z^{tgt}_t)$, where $z^{ref}_t$ and $z^{src}_t$ are extracted directly from the sequences $\{z^{src}_T\}^T_{t=1}$ and $\{z^{ref}_T\}^T_{t=1}$ respectively, and $z^{tgt}_t$ is derived from the denoised $z^{tgt}_{t+1}$ of the preceding step. Denote $h^{ref}_i, h^{src}_i, h^{tgt}_i$ as the features prior to entry into the $i$-th self-attention layer. These features are independently transformed into query, key, and value components via the following projections:
\begin{align}
Q_r, K_r, V_r &= W_Q h^{ref}_i, W_K h^{ref}_i, W_V h^{ref}_i, \notag \\
Q_s, K_s, V_s &= W_Q h^{src}_i, W_K h^{src}_i, W_V h^{src}_i, \\
Q_t, K_t, V_t &= W_Q h^{tgt}_i, W_K h^{tgt}_i, W_V h^{tgt}_i, \notag
\end{align}
where $W_Q, W_K, W_V$ represent three distinct projection matrices. The original self-attention mechanism operates as follows:
\begin{equation}
Attn(Q,K,V) = \text{Softmax}(\frac{QK^T}{\sqrt{d}}) \cdot V,
\end{equation}
where $d$ denotes the dimensionality of the key and query features. In our approach, we substitute the self-attention operation in $Q_t, K_t, V_t$ with the Style Attention Control. More specifically, we replace the original key and value features $K_t, V_t$ with $K_r, V_r$ from reference images and execute a dual-branch cross-attention mechanism:
\begin{align}
    O_t = Attn(Q_t,K_r,V_r), \notag \\
    O_s = Attn(Q_s,K_r,V_r).
\end{align}
Importantly, in this framework, $Q_t$ acquires a more pronounced style after undergoing multiple denoising iterations. In contrast, $Q_s$, being derived directly from $z_t^{src}$, preserves greater fidelity to the source information. Consequently, the style information in the output $Q_s$ is comparatively subdued. To control the stylization intensity, we introduce the Style Guidance technique for the fusion of $Q_s$ and $Q_t$:
\begin{equation}
    O_t = O_s + \omega(O_t - O_s),
\end{equation}
where $\omega$ signifies the Style Guidance scale. When $\omega>1$, Style Guidance accentuates the differences between $Q_s$ and $Q_t$, thereby yielding outputs with a stronger style.

Furthermore, the number of denoising steps and the number of self-attention layers implementing Style Attention Control significantly affect the balance between source preservation and stylization. As the findings in \cite{Cao_2023_MasaCtrl}, it is noted that the query features within the encoder part of the U-Net often exhibit a deficiency in layout and structural information. Consequently, Style Attention Control is selectively executed only within the median and decoder parts of the U-Net. Regarding the adjustment of denoising steps, we initially assign a null value to the Style Guidance weight in the preliminary denoising phase, encapsulated by the following formulation:
\begin{equation}
    \omega=\left\{\begin{array}{l}
0, \quad \text { if } t<S, \\
\omega, \quad \text { otherwise, } 
\end{array}\right.
\end{equation}
where $S$ represents the timestep indicates the beginning of applying Style Guidance. Setting $S$ to $0$ implies that the stylization outcomes are predominantly influenced by the current source latent code $z^{src}_t$, thus retaining a substantial portion of the original source content and manifesting a muted style.

\begin{figure}[h]
\begin{center}
\includegraphics[width=0.9\linewidth]{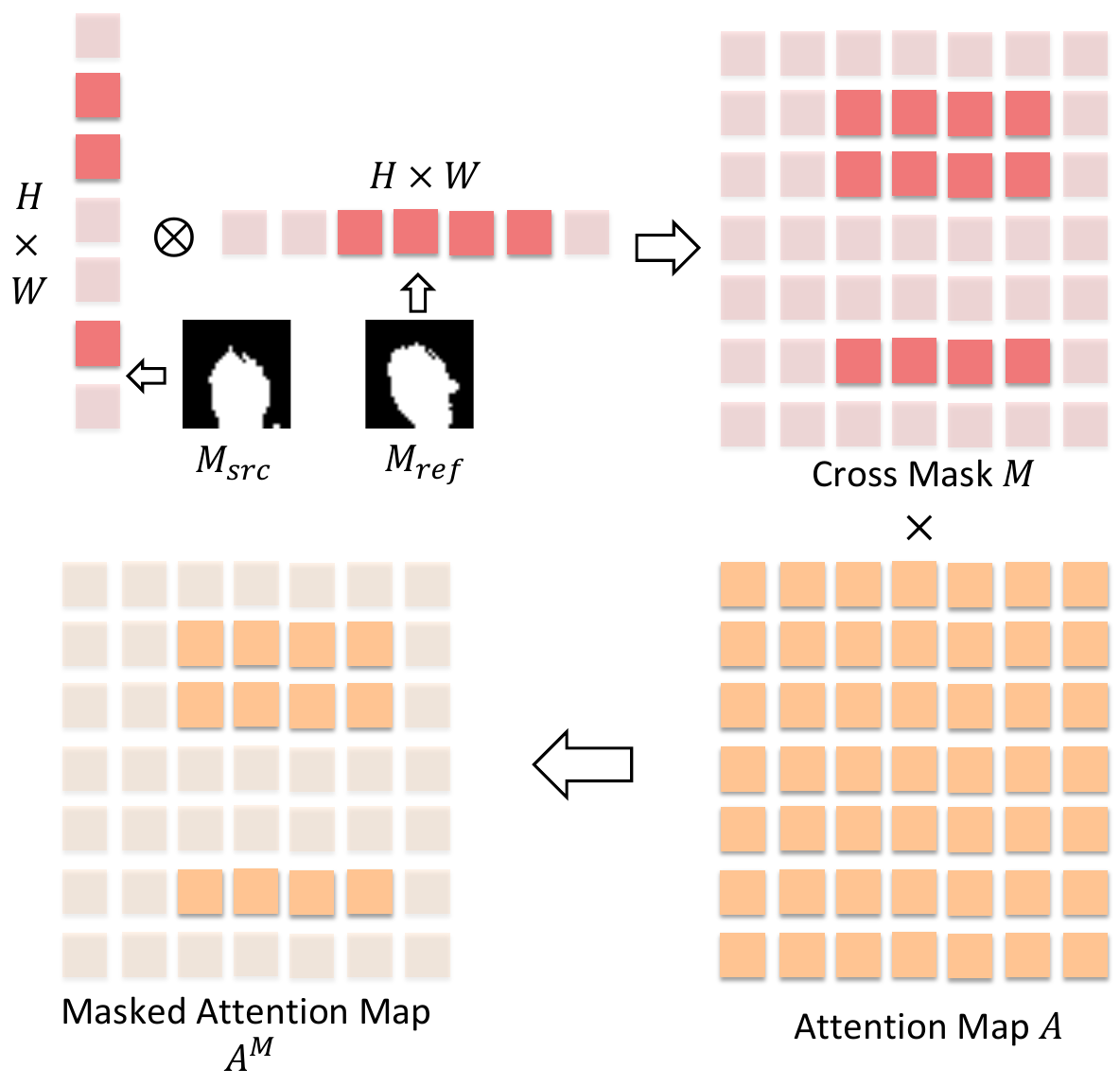}
\end{center}

   \caption{Illustration of Cross Mask for Attention Map.}
\label{fig:crossmask}
\end{figure}

\noindent
\textbf{Mask Prompted Style Attention Control.} Direct queries of features within the attention space may may lead to query confusion problems, particularly when dealing with images set against intricate backgrounds. To address this issue, we further propose a Mask Prompted Style Attention Control. This technique, diverging from the querying of features across all regions, specifically targets features within masked areas that share identical facial attributes. A dedicated face attributes segmentation model can be employed for precise facial mask extraction, such as Segment Anything \cite{kirillov2023segment}. We represent the extracted facial masks from the source and reference images as $M_{s}$ and $M_{r}$, respectively.

The original masking technique in self-attention is constrained to applying a single mask to the attention map, rendering it inadequate for our Style Attention Control. To circumvent this limitation, we introduce a cross-masking approach. Given that our attention map is computed based on the similarity between queries from the source and keys from the reference, each axis of the attention map can distinctly represent the source and the reference. As illustrated in Fig. \ref{fig:crossmask}, this allows for the application of masking on the attention map in two dimensions using two separately flattened masks. 
Specifically, for an attention map $A = \text{Softmax}(\frac{Q_t K_r^T}{\sqrt{d}}) \in \mathbb{R}^{H\times L\times L}$ 
within the Style Attention Control, we initially flatten the masks $M_{s} \in \mathbb{R}^{H\times W}$ and $M_{r} \in \mathbb{R}^{H\times W}$ into $M_{s}, M_{r} \in \mathbb{R}^{1\times L}$, respectively. 
Subsequently, the Cross Mask $M = M_{s} \cdot M_{r}^T$ is derived by matrix multiplication of $M_{s}, M_{r}$, 
and then applied to obtain the masked attention map by 
$A^M = A \times M_{s}$
enabling the execution of cross mask attention as follows:
\begin{align}
Attn(Q,K,V,M_{s},M_{r}) = A^M \cdot V.
\end{align}
The formulation of Mask Prompted Style Attention Control is articulated as follows:
\begin{align}
O^M_s &= Attn(Q_s,K_r,V_r,M_{s},M_{r}), \notag \\
O^M_t &= Attn(Q_t,K_r,V_r,M_{s},M_{r}), \notag \\
O^M_t &= O^M_s + \omega(O^M_t - O^M_s).
\label{eq:maskprompting}
\end{align}
For the background regions, Mask Prompted Style Attention Control is executed utilizing the inverse background masks $1-M_{s}$ and $1-M_{r}$. The outcome of implementing Eq. \ref{eq:maskprompting} with $1-M_{s}$ and $1-M_{r}$ is denoted as $O^{M^{-1}}_t$. Subsequently, the final output $O_t$ is computed by:
\begin{equation}
O_t = M_{s}\otimes O^M_t + (1-M_{s}) \otimes O^{M^{-1}}_t.
\label{eq:backgroundmix}
\end{equation}

\noindent
\textbf{Chain-of-Painting.} Merely isolating a face image from the foreground and background proves insufficient to address the scenario where different facial attributes exhibit similar color patterns. Consequently, we introduce Chain-of-Painting, a face stylization paradigm influenced by the sequential manner in which humans paint a face, progressing from coarse to detailed representation.
Chain-of-Painting, rooted in our Mask Prompted Style Attention Control, employs a sequence of facial attribute masks, derived via a face parsing model, to augment the stylized results on a per-attribute basis. Fig. \ref{fig:chain-of-painting} demonstrates the Chain-of-Painting process. Specifically, a user may begin by synthesizing a result without employing Mask Prompting, and then proceed to iteratively redraw any regions identified as unsatisfactory through an interactive process. In each step of the Chain-of-Painting process, the stylized outcome supersedes the source image from the preceding phase, accompanied by a corresponding mask informed by human feedback. Some example results are shown in Fig. \ref{fig:cop} (a).

This requirement mandates the maintenance of unmasked regions in their previous state. To accomplish this, we propose replacing Eq. \ref{eq:backgroundmix} with
\begin{align}
O_t &= M_s \otimes O_t + (1-M_s) \otimes O_s,
\end{align}
where $O_s$ signifies the output of the self-attention mechanism deployed on the source feature, namely, $O_s = Attn(Q_s, K_s, V_s)$. Consequently, the unmasked region is replenished with the source images.

\begin{figure*}[!t]
\begin{center}
\includegraphics[width=0.95\linewidth]{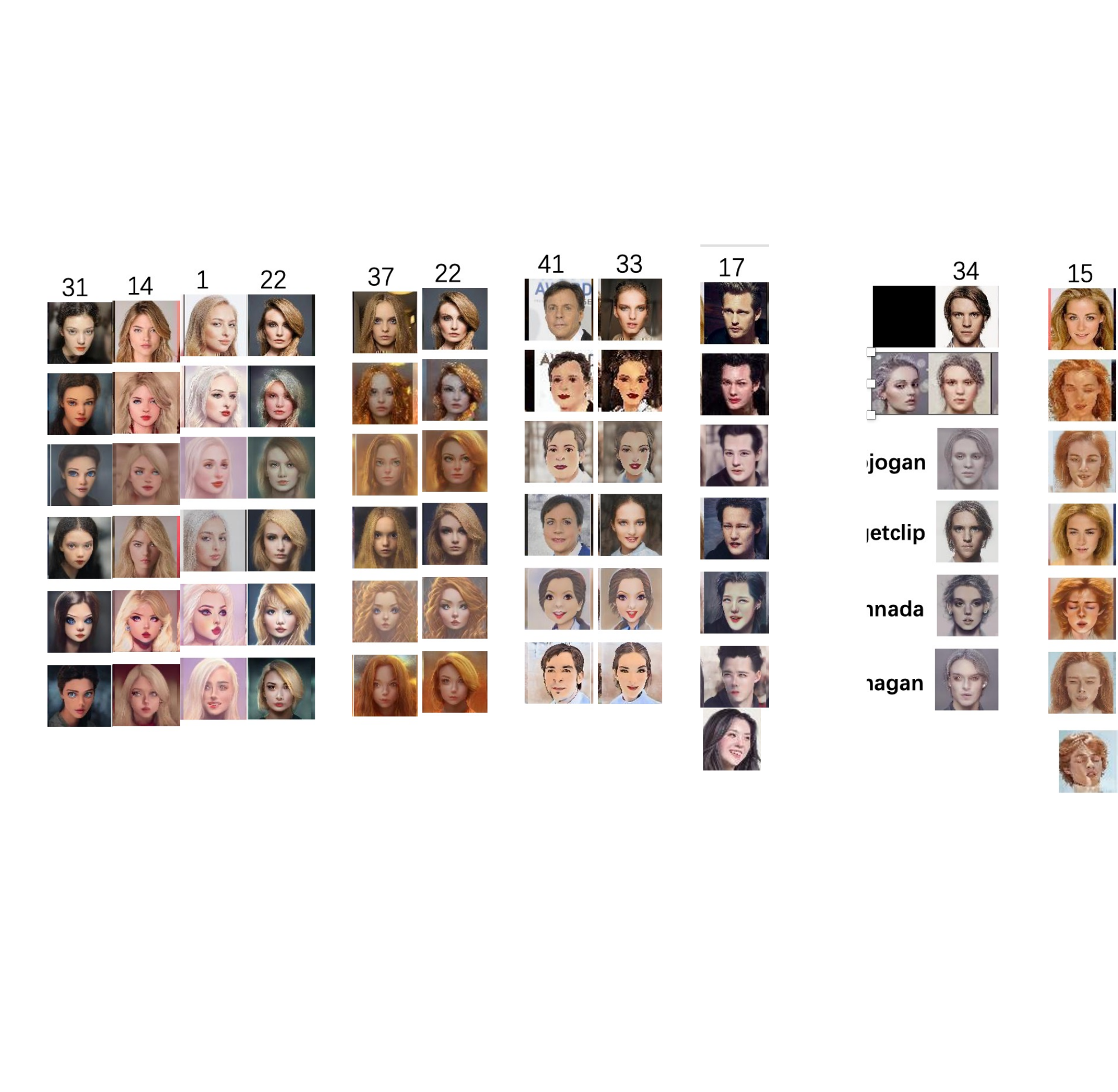}
\end{center}

   \caption{Qualitative comparison with the state-of-the-art methods.}
   
\label{fig:final_comp}
\end{figure*}

In addition to enhancing a stylized result, Chain-of-Painting can be employed to synthesize images using multiple style references. Rather than utilizing a singular reference image in each stage of the Chain-of-Painting, it is possible to introduce a distinct reference image during the current phase of the process and delineate the target area for stylization using a mask as the results shown in Fig. \ref{fig:cop} (b).

\section{Experiments}

\noindent
\textbf{Implementation Details.} Our Portrait Diffusion framework is built upon Stable Diffusion v1.5 \footnote{https://huggingface.co/runwayml/stable-diffusion-v1-5}, a version of Latent Diffusion Models (LDMs) \cite{rombach2022high} that has been pre-trained on the extensive LAION-5B text-image dataset \cite{schuhmann2022laion}. We utilize a fixed timestep of 50 for both the process of converting images into latent codes using the DDIM deterministic inversion method \cite{song2020denoising}, and for generating images from noise using the DDIM sampling strategy \cite{song2020denoising}. The conditional text prompt for Stable Diffusion is configured as 'head'. Moreover, the classifier-free guidance scale is designated as 0. For the implementation of Style Guidance, we assign $\omega$ a value of $1.2$ and set $S$ at $35$ as default. Our experimental procedures are executed on a solitary NVIDIA 4090 GPU. Reference images are sourced mainly from the AFHQ dataset \cite{liu2021blendgan}, while all content images are derived from the CelebA-HQ dataset \cite{karras2018progressive}. The resolution for all images is set to 512 × 512 pixels.

\subsection{Comparison with SOTA methods}
Our comparison experiments mainly focus on contrasting the proposed Portrait Diffusion with the current state-of-the-art (SOTA) one-shot adaptation methods. This includes StyleGAN-based approaches such as JoJoGAN \cite{chong2022jojogan}, TargetCLIP \cite{chefer2022image}, StyleGAN NADA (NADA) \cite{gal2022stylegan} and DynaGAN \cite{kim2022dynagan}, in addition to diffusion-based methods like InST \cite{zhang2023inversion} and VCT \cite{cheng2023general}. All stylization results are produced by their corresponding open-source codes.

\noindent
\textbf{Qualitative Comparison.}
Fig. \ref{fig:final_comp} presents a qualitative comparison of the comparison results. The first column displays the original natural faces, followed by the reference artistic portraits in the second column. The results of our Portrait Diffusion are showcased in the third column, with the remaining columns featuring outputs from various other models. In the case of StyleGAN-based methods, as depicted in Fig. \ref{fig:final_comp}, JoJoGAN demonstrates suitable style variations. However, it introduces unintended semantic feature changes compared to the original content, particularly noticeable in the eyes (first row) and posture (third row). TargetCLIP's outputs, on the other hand, suffer from poor content consistency and fail to replicate the reference image's style. Both NADA and DynaGAN successfully capture the reference style but alter facial features significantly. Among the diffusion-based methods, InST struggles with overfitting, leading to some suboptimal results. VCT strikes a better balance between maintaining original content and achieving style transfer. Nevertheless, in more challenging cases (first and last rows), its outputs exhibit considerable alterations in expression. Differing from these methods, Portrait Diffusion not only achieves varied styles but also excels in preserving local details such as facial features and hair texture, thus maintaining a consistent facial identity between source and output images. Additionally, our method adeptly captures fine-grained styles from the reference image, evident in the eyeshadow and hair highlights in the third row and the detailed hair texture in the fifth row.

\begin{figure}[!t]
\begin{center}
\includegraphics[width=1\linewidth]{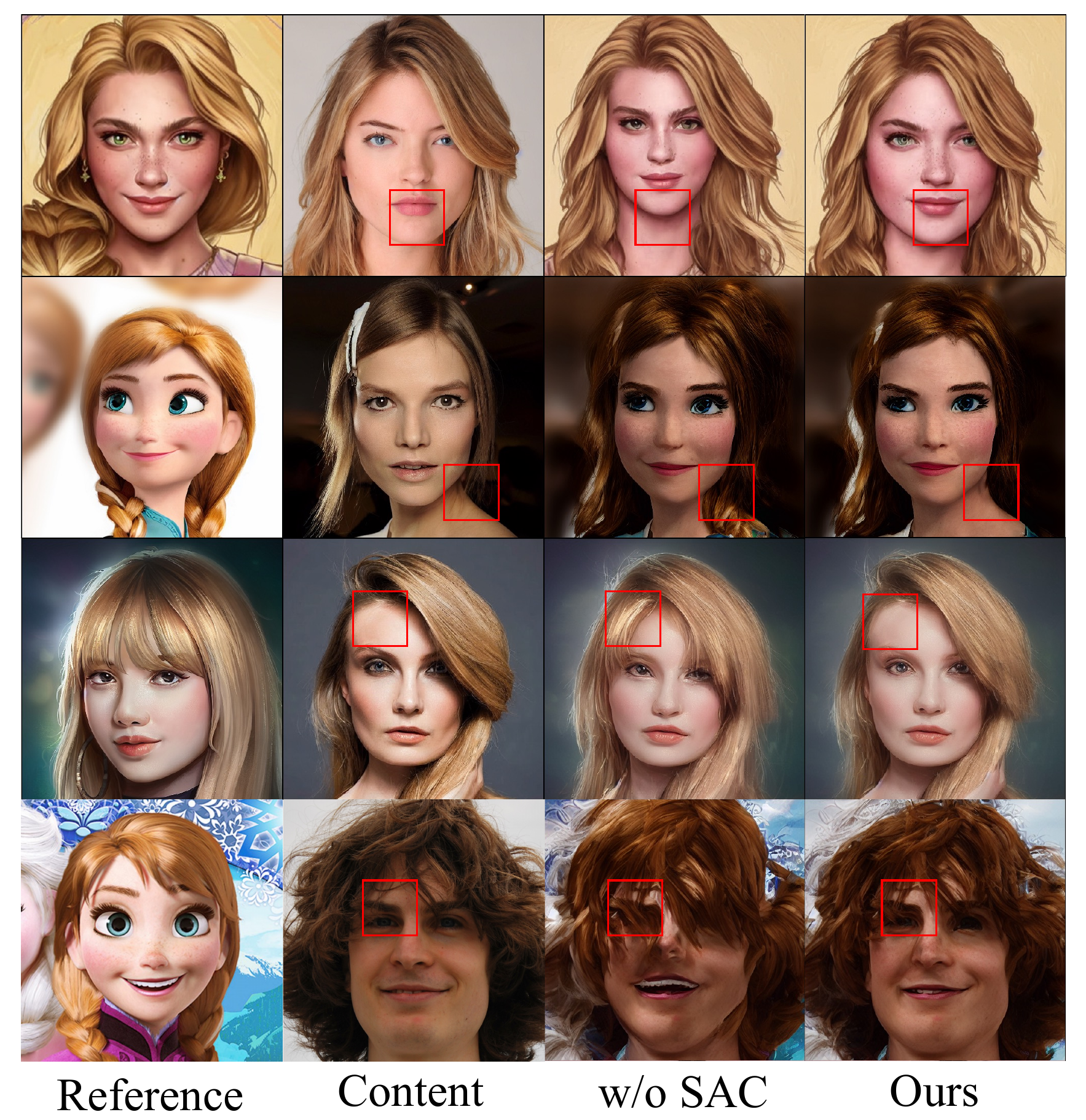}
\end{center}
\caption{Ablation results on the methods of attention control, where SAC stands Style Attention Control.}
\label{fig:abla_controls}
\end{figure}






\noindent
\textbf{Quantitative Comparison.} The task of portrait stylization inherently lacks an objective ground truth for result evaluation, rendering it a predominantly subjective domain. Consequently, we utilize user studies as a means to discern which methods yield the most human-preferred outcomes. For each model, we collected responses for 20 distinct source-target pairings, amassing a total of 20 answers. As indicated in Table \ref{tab:pref_times}, more than 30\% of the results from our proposed Portrait Diffusion method were identified as the most favored results. This significant preference underscores the effectiveness and appeal of Portrait Diffusion in the context of portrait stylization.

We further evaluated the time efficiency of generating a single stylized image. As detailed in Table \ref{tab:pref_times}, Portrait Diffusion stands out as a zero-shot technique, obviating the need for model fine-tuning and thereby saving time. In contrast, the fastest methods such as JoJoGAN also require at least 48 seconds for fine-tuning, while other methods take from hundreds to thousands of seconds. Considering that learning a good style representation may require multiple fine-tuning sessions, this results in a considerable time expense, limiting the practical application of these methods.

\begin{table}[ht]

\centering
\begin{tabular}{c|cc}
\hline
Model  & Rate & Fine-tuning Times   \\ \hline
 JoJoGAN \cite{chong2022jojogan}    &    \underline{20.52\%} &      \underline{48.52s}      \\
  TargetCLIP \cite{chefer2022image}  &      15.79\%    &      692.48s        \\
NADA\cite{gal2022stylegan}  &     8.94\%    &     155.32s          \\
NydaGAN \cite{kim2022dynagan}   &      12.89\%  &      1156.815s        \\
InST \cite{zhang2023inversion}   &     2.89\%  &     2007.96s            \\
VCT \cite{cheng2023general}   &     8.68\%  &     374.11s          \\
\hline
\textbf{Ours} & \textbf{30.26\%}   & \textbf{0s}           \\ \hline
\end{tabular}
\caption{Quantitative comparison on user preference rate and the time consumption.}\label{tab:pref_times}
\end{table}

\subsection{Ablation Study}

We conducted visual ablation studies to evaluate the impact of the attention control method in portrait stylization. Figure \ref{fig:abla_controls} illustrates that synthesized images lacking Style Attention Control undergo significant semantic changes and exhibit artifacts. In contrast, our method, while introducing style variations, effectively preserves the semantic integrity of the content images. Further, the incorporation of the Chain-of-Painting technique enhances this consistency, aligning closely with the source material without diminishing the style elements. This demonstrates the efficacy of our approach in balancing stylistic modifications with content accuracy.

\section{Conclusion}

In this study, we propose Portrait Diffusion, a novel training-free framework for face portrait stylization. Portrait Diffusion can synthesize stylized portraits in a zero-shot manner and finely control the style. The proposed Style Attention Control can effectively fuse the features of content and reference images in each denoising iteration. Furthermore, we introduce Chain-of-Painting, which not only refines the less satisfactory stylized regions from coarse to fine but also achieves portrait stylization utilizing multiple style references. Extensive experimental results demonstrate that our method is capable of synthesizing high-quality stylization results while preserving the content of the source image, distinctly outperforming existing state-of-the-art methods.

\newpage
{
    \small
    \bibliographystyle{ieeenat_fullname}
    \bibliography{main}
}

\clearpage
\setcounter{page}{1}
\setcounter{figure}{0}
\setcounter{table}{0}

\appendix

\section*{Appendix}
In this appendix, additional details and analyses of our approach are provided. The detailed implementation algorithm is presented in Appendix \ref{appendix:imple}.  Appendix \ref{appendix:quan} offers additional quantitative comparison results. Furthermore, more ablation styles and visual results are provided in Appendices \ref{appendix:abla} and \ref{appendix:result}, respectively. Limitations and Social Impacts are discussed in Appendix \ref{appendix:limit}.

\section{More Implementation Details}
\label{appendix:imple}

\begin{figure*}[th]
\begin{center}


\subfloat[]{
\includegraphics[width=1\linewidth]{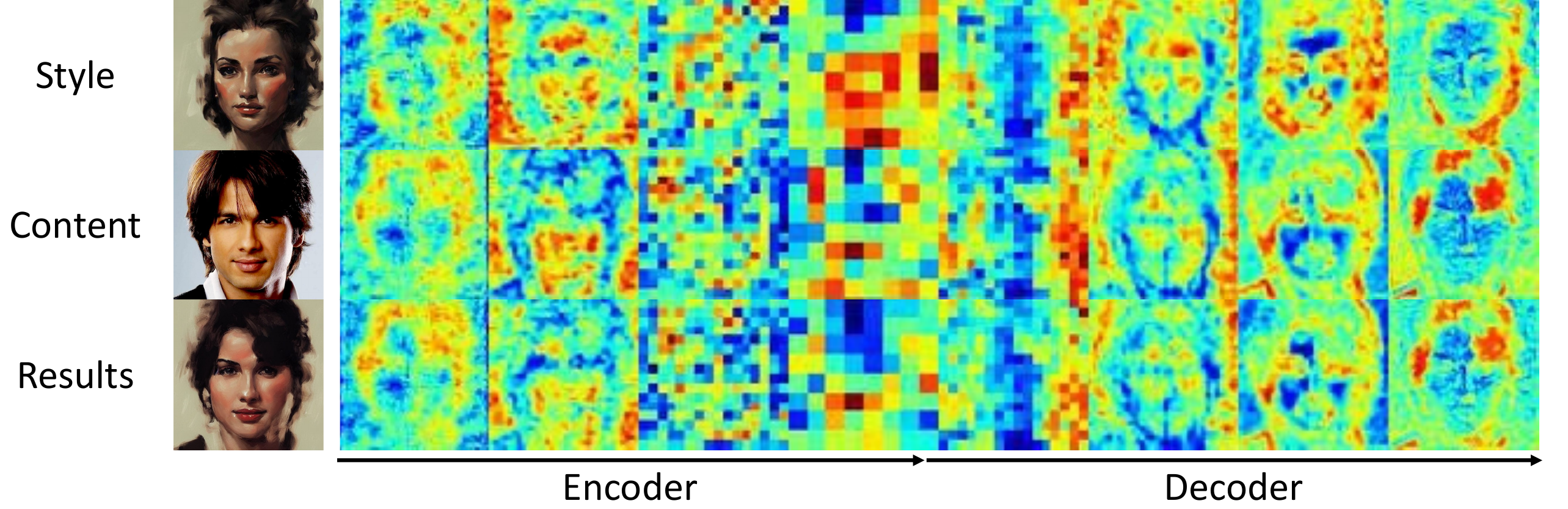}
}
\
\subfloat[]{
\includegraphics[width=1\linewidth]{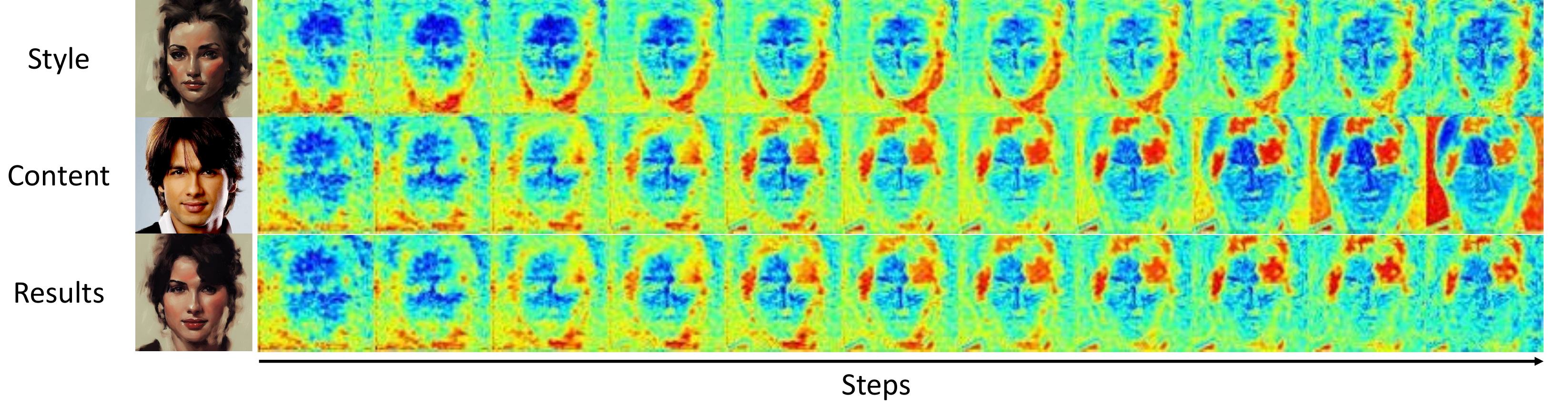}}
\end{center}

   \caption{Visualization of the projected Query features. (a) Different self-attention layers at 35 timesteps. (b) Different timesteps in the 16th self-attention layers. }
\label{fig:attn_q}
\end{figure*}

The complete algorithm for Portrait Diffusion is delineated in Algorithm \ref{alg::pdiff}. Empirically, it was found that a minimum of 50 steps is required for DDIM inversion; fewer steps lead to distorted results. Regarding the Style guidance scale, it is set to $1.2$ to strike a balance between style transfer and content preservation.

In Fig. \ref{fig:attn_q}, we visualize the self-attention of different layers and time steps in the U-Net network, focusing on the Query features of content, style, and result images, to identify the optimal timing for Style Attention Control (SAC). In Fig. \ref{fig:attn_q} (a), within the Encoder part, the query features appear somewhat disordered, whereas, in the Decoder part, they reveal clearer facial contours. In Fig. \ref{fig:attn_q} (b), the earlier time steps display only rudimentary facial contours, which become progressively more defined as the time steps increase. Consequently, for layer selection, we opt to commence SAC from the Decoder section, i.e., the $10$th self-attention layers in U-Net. Regarding the SAC step $S$, we begin SAC from step 35 (out of a total of 50 time steps).

\begin{algorithm}[h] 
	\caption{Portrait Diffusion} 
	\label{alg::pdiff}
	\begin{algorithmic}[1] 
		\Require 
		$x^{src}$, $x^{ref}$: source and reference image.
            \Require 
            $\epsilon_\theta$: Pre-trained diffusion model. $T$: Time-step, $P$: Prompt, $\omega$: Style guidance sacle, $S$: SAC step.
            \State Compute  $\{z^{src}_T\}^T_{t=1},\{z^{ref}_T\}^T_{t=1}$ with DDIMEncode.
            \State $z^{tgt}_T=z^{src}_T$
		\For{$t = T, T-1, ..., 1$};
                \State $z^{src}_t,z^{ref}_t  \leftarrow \{z^{src}_T\}^T_{t=1},\{z^{ref}_T\}^T_{t=1}$
                \State $ \{Q_s, K_s, V_s\}\leftarrow \epsilon_\theta(z^{src}, P, t)$;
                \State $\{Q_r, K_r, V_r\}\leftarrow \epsilon_\theta(z^{ref}, P, t)$;
                \State $\{Q_t, K_t, V_t\} \leftarrow \epsilon_\theta(z^{tgt}_t,P,t)$;
                \If{$t>=S$}
                \State $\epsilon = \epsilon_\theta(z_t, P, t; \text{SAC}(\{Q_t, Q_s, K_r, V_r;\omega\}))$;
                \Else
                \State $\epsilon = \epsilon_\theta(z_t, P, t; \text{SAC}(\{Q_t, Q_s, K_r, V_r;0\}))$;
                \EndIf
                \State $z^{tgt}_{t-1} \leftarrow \text{Sample}(z^{tgt}_{t}, \epsilon)$;
  
            \EndFor
        \State Compute target image $x^{tgt}=\mathcal{D}(z^{tgt}_{0})$ with decoder $\mathcal{D}$; 
    
	\end{algorithmic} 
 \label{algorithm}
\end{algorithm}

\section{More Quantitative Comparisons}
\label{appendix:quan}

\begin{table}[ht]
\centering
\begin{tabular}{c|cc}
\hline
Model & ID $\uparrow$ & Style $\downarrow$     \\ \hline

NADA \cite{gal2022stylegan} & 0.252 & {1.101} \\
TargetCLIP \cite{chefer2022image} & \underline{0.554} & {1.50} \\
JoJoGAN \cite{chong2022jojogan} & {0.351} & 1.169 \\
DynaGAN \cite{kim2022dynagan} & 0.164 & \textbf{0.667} \\
InST \cite{zhang2023inversion} & 0.067 & 9.819 \\
VCT \cite{cheng2023general} & 0.381 & 1.228 \\ \hline
\textbf{Ours} & \textbf{0.581} & \underline{0.722} \\
\hline


\end{tabular}

\caption{Quantitative comparison on Face Identity Similarity and Image Style Loss. The best and second best metrics are emphasized in \textbf{boldface} and \underline{underline} respectively. ($\downarrow$) signifies that a lower value is preferable, while ($\uparrow$) denotes that a higher value is preferable.}\label{tab_losses}
\end{table}

To further demonstrate the superiority of our method, we employ two quantitative metrics for evaluating the results generated by various comparative methods: Image Style Loss (Style) and Face Identity Similarity (ID). Image Style Loss is calculated through features extracted by a pre-trained VGG network \cite{huang2017arbitrary}, assessing the style transfer capabilities of the model by measuring the style discrepancy between generated images and reference images. Face Identity Similarity, on the other hand, utilizes a pre-trained facial recognition network, Arcface \cite{deng2019arcface}, to compute the similarity in facial identity between generated images and content images, thereby gauging the model's ability to preserve the content information of the source image. Specifically, we tasked all comparative methods with generating 580 distinct stylized facial images based on 580 pairs of content and style images. Subsequently, we computed the Style and ID metrics for each comparative method. The quantitative comparison results are presented in Tab.\;\ref{tab_losses}. It is evident that our method not only possesses robust style transfer capabilities (as indicated by a low Style loss of 0.722) but also excels in preserving content information, achieving an average ID similarity of 0.581. However, it is important to note that evaluating portrait stylization is a highly subjective task. Quantitative metrics may not fully capture the aesthetic value of generated images, instead providing a limited analysis of the alignment between the generated outcomes and their content and style.

\section{More Ablation Study}
\label{appendix:abla}

\begin{figure*}[th]
\begin{center}


\subfloat[]{
\includegraphics[width=1\linewidth]{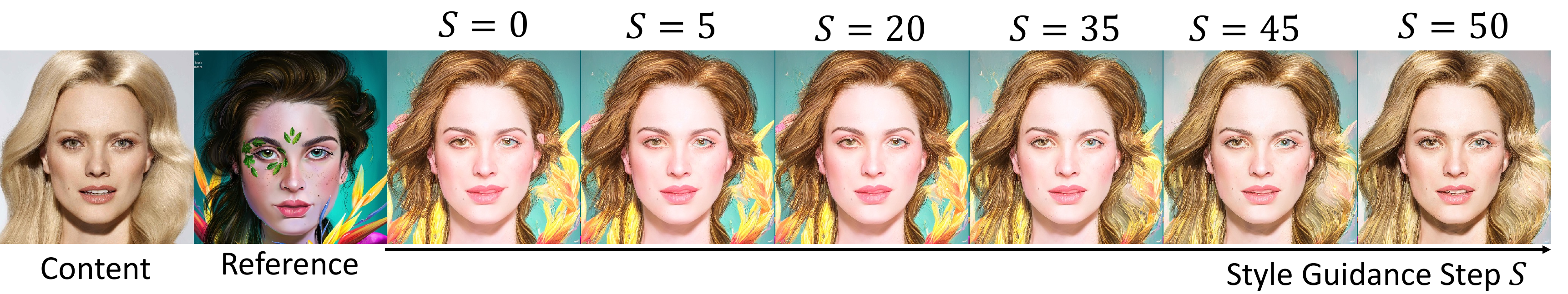}}
\
\subfloat[]{
\includegraphics[width=1\linewidth]{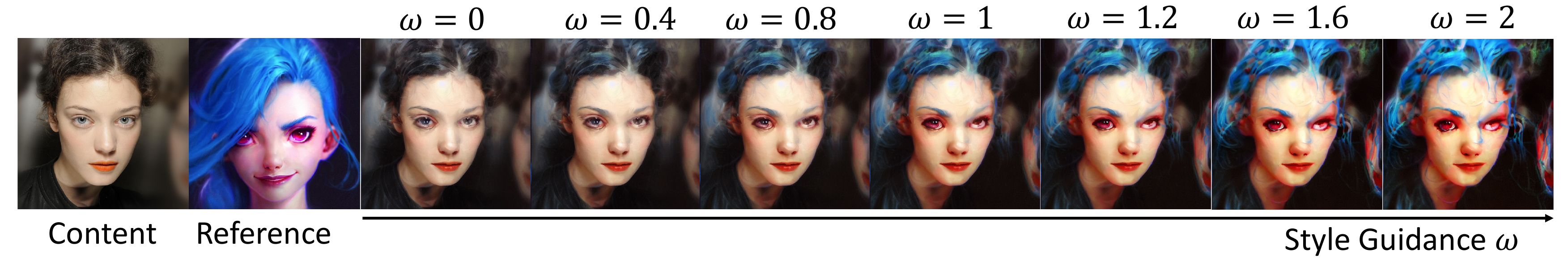}}

\subfloat[]{
\includegraphics[width=1\linewidth]{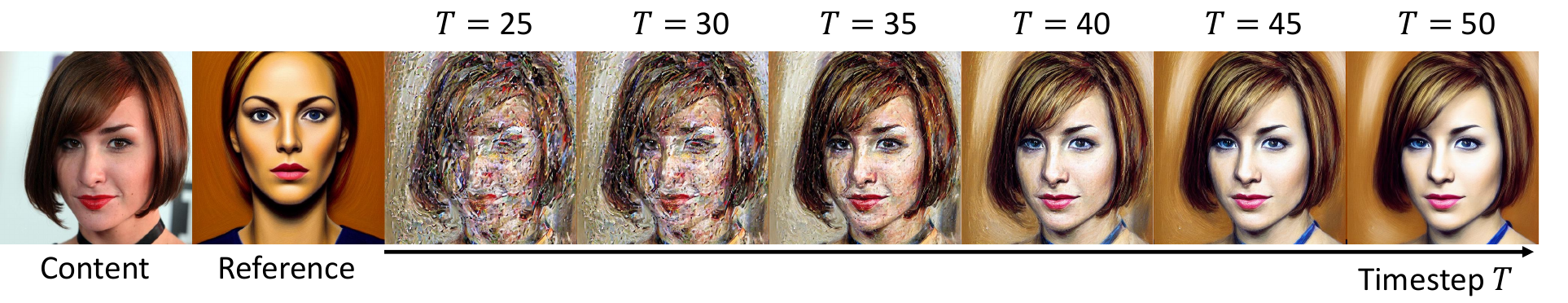}}
\end{center}
\vspace{-15pt}
   \caption{Ablation results on Style Guidance. (a)Results of different Style Guidance Scale. (b) Results of applying Style Guidance starting from different denoising steps. (C) Results with difference time-step $T$.}
\label{appendixfig:abla_guidance}
\end{figure*}

We conducted a further analysis to evaluate the efficacy of Style Guidance. The results depicted in Fig. \ref{appendixfig:abla_guidance} (a) illustrate that varying the style guidance scale in the style attention control mechanism markedly influences the final synthesized outcomes. An increase in the style guidance weight progressively amplifies the style elements in the synthesized images. When the weight $\omega$ is less than 0.8, the resultant images exhibit minimal style influence. Conversely, weights exceeding 1.5 lead to the emergence of undesirable textures. Figure \ref{fig:abla_guidance} (b) displays the outcomes of applying Style Guidance at different timesteps. These results indicate that lower timesteps $S$ produce outputs with substantial deviations from the original content. In contrast, as $S$ increases, the outputs demonstrate enhanced content fidelity, albeit with a corresponding reduction in stylistic attributes.


\section{More Results of Portrait Stylization}
\label{appendix:result}

\begin{figure*}[t]
\begin{center}
\includegraphics[width=\linewidth]{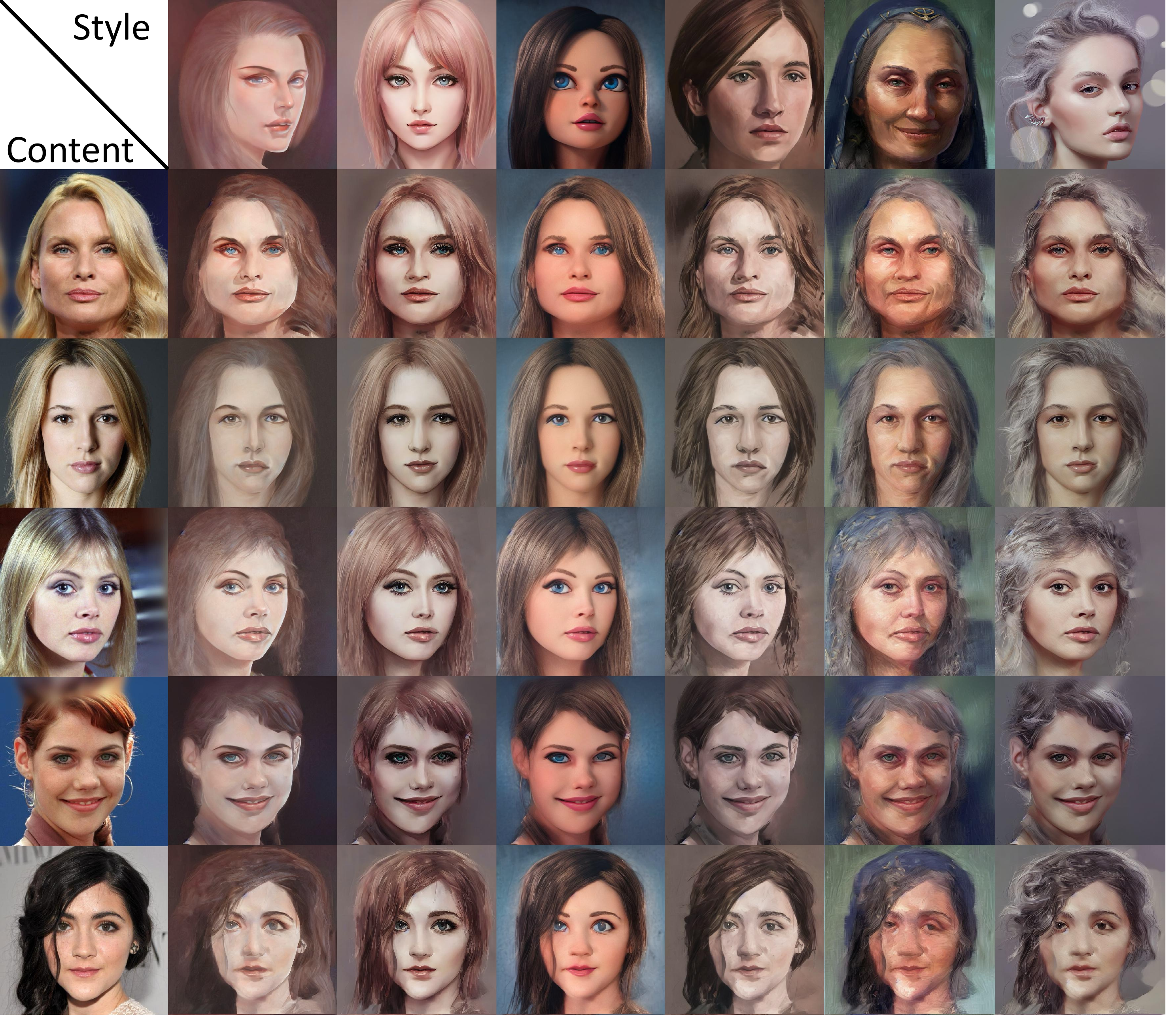}
\end{center}
\caption{ More synthesis results. The first column contains the reference images, and the first row contains the content images. The other images are the model outputs based on corresponding content and reference images.
}
\label{appendixfig:show_case_1}
\end{figure*}

To further verify the model performance in the Portrait Stylization task, we conducted additional experiments using a variety of content images and reference styles, as depicted in Fig.\;\ref{appendixfig:show_case_1}. The figure displays the reference images in the first row and the content images in the first column. The model outputs, as observed in the subsequent cells, demonstrate the method's exceptional efficacy in preserving content while effectively transferring style.

\section{Limitations and Social Impacts}
\label{appendix:limit}
\noindent
\textbf{Limitations.} Despite the fact that Portrait Diffusion is a training-free method for portrait stylization, its main limitation, similar to other works based on the diffusion model, lies in the need for multiple function evaluations (NFEs) during image generation. By adopting DDIM sampling for generating images, we only require 50 NFEs, reducing the time for generating a single image to just 11 seconds using an NVIDIA 4090 GPU. However, compared to methods based on Generative Adversarial Networks (GANs), which require only one NFE, there is a significantly greater time expenditure in image generation.

\noindent
\textbf{Social Impacts.} Our method facilitates efficiency enhancement and inspiration for professionals, while also allowing novices to effortlessly generate facial portraits in their preferred styles. We aspire for our approach to broaden public engagement with the appreciation and creation of artistic portraits. On the other hand, we hope that this technology is utilized responsibly, refraining from creating unauthorized images and eschewing the production of art with offensive content.

\clearpage

\end{document}